\documentclass[conference]{IEEEtran}
\IEEEoverridecommandlockouts
\usepackage{cite}
\usepackage{amsmath,amssymb,amsfonts}
\usepackage{algorithmic}
\usepackage{graphicx}
\usepackage{textcomp}
\usepackage{xcolor}
\usepackage{booktabs}
\usepackage{multirow}
\usepackage{array}
\usepackage{hyperref}
\usepackage{subcaption}

\def\BibTeX{{\rm B\kern-.05em{\sc i\kern-.025em b}\kern-.08em
    T\kern-.1667em\lower.7ex\hbox{E}\kern-.125emX}}

\begin{document}

\title{\huge Bio-PM-Guided Synthetic IMU Generation}
\author{
    \IEEEauthorblockN{Deepali Singh, Deepesh Parpani, Aviral Gupta}
    \IEEEauthorblockA{
        College of Information and Computer Sciences \\
        University of Massachusetts Amherst \\
        Amherst, USA \\
        \{deepalisingh, dparpani, aviralgupta\}@umass.edu\\
    }
}

\maketitle

\begin{abstract}

Collecting and labeling inertial measurement unit (IMU) data for human activity recognition (HAR) is expensive, and models trained on one group of subjects often fail to generalize to unseen individuals. Even with balanced data, classifiers struggle to distinguish dynamically similar activity classes and benefit from targeted augmentation. We explore whether synthetic IMU data generated in the representation space of a pretrained biosignal encoder, Bio-PM \cite{b1}, can serve as a practical augmentation source. Rather than synthesizing raw signals, we leverage Bio-PM's frozen encoder to convert 10-second IMU windows into token embedding sequences and train class-conditional decoders in this space: (1) an autoregressive transformer decoder that predicts tokens sequentially, and (2) a diffusion decoder that learns the token distribution through a denoising objective. We analyze generation realism through distributional metrics and embedding-space visualizations, evaluate downstream utility through leave-one-subject-out HAR classification, and present a PatchDecoder for approximate raw waveform reconstruction. Our diffusion decoder achieves a Train-on-Synthetic-Test-on-Real (TSTR) macro-F1 of 0.68, demonstrating that generated tokens are realistic enough to train a classifier that transfers to real data. Combining synthetic-token augmentation with a 320-dimensional gravity-aware feature representation raises LOSO macro-F1 from 0.729 to 0.809 over a real-only baseline across 51 subjects on the WISDM benchmark, with statistically significant gains concentrated in the most challenging activity classes. These results demonstrate that high-fidelity generation in a pretrained encoder's representation space is achievable and practically useful for cross-subject HAR augmentation. Source Code is available here: \href{https://drive.google.com/drive/folders/1ZuyGTtrXoHTMf8kGgOoRIvQO4KELXhd_?usp=sharing}{Google Drive Link}
\end{abstract}


\section{Introduction}

Human activity recognition (HAR) from wrist-worn inertial measurement unit (IMU) sensors supports a range of health monitoring and behavioral analysis applications. Despite strong benchmark performance, cross-subject generalization remains a persistent challenge as classifiers trained on one subject population frequently fail on unseen individuals, a gap directly measured by leave-one-subject-out (LOSO) evaluation.

Synthetic data augmentation offers one path toward broader generalization, yet generating realistic IMU signals directly is difficult given complex temporal structure, subject-specific motion styles, and device-placement variability. An alternative is to generate in a pretrained model's representation space, where activity-relevant structure is already encoded. Bio-PM \cite{b1}, a Transformer encoder pretrained via masked movement-segment reconstruction on the NHANES wrist-IMU corpus (approximately 28,000 hours, 11,000 participants), provides this foundation. Its token space is semantically organized by activity and exhibits no subject-level clustering, making it a natural target for class-conditional generation without subject conditioning.

We train class-conditional generative decoders directly in Bio-PM's token space, using the frozen encoder as a quality oracle. A key challenge is that Bio-PM produces contextual, BERT-like representations in which each of the 192 tokens per window encodes global temporal context, making this a richer but more complex distribution to model than a standard latent space. We develop and compare two decoders: (1) CAR-IMU, a Class-conditional AutoRegressive decoder, and (2) CD-IMU, a Class-conditional Diffusion decoder, evaluating both distributional realism and downstream LOSO classification utility on the WISDM benchmark.

The main contributions of this paper are as follows:
\begin{itemize}
    \item We propose CAR-IMU, which generates synthetic Bio-PM token sequences autoregressively, and characterize a distributional shift requiring per-class affine calibration to resolve.
    \item We design CD-IMU, a diffusion decoder that eliminates the calibration dependency of CAR-IMU, achieving high distributional realism without post-hoc correction.
    \item We provide systematic generation quality analysis across both decoders using cosine similarity, Fr\'echet distance (FD) in Bio-PM token space, maximum mean discrepancy (MMD), variance ratio, and Train-on-Synthetic-Test-on-Real (TSTR) evaluation.
    \item We demonstrate that augmenting a LOSO HAR classifier with CD-IMU synthetic tokens, using a 320-dimensional feature representation and confusion-aware augmentation, yields statistically significant improvements over the real-only baseline across 51 subjects on WISDM.
    \item We present a PatchDecoder for approximate reconstruction of raw IMU waveforms from synthetic token sequences, extending the pipeline toward signal-level synthesis.
\end{itemize}

\section{Dataset and Bio-PM}

\subsection{WISDM Dataset}
We use the WISDM wrist accelerometer dataset \cite{b2}, comprising recordings from 51 subjects performing six activities: Walking, Jogging, Upstairs, Downstairs, Sitting, and Standing. Raw signals are sampled at 20 Hz and segmented into non-overlapping 10-second windows, yielding 12,514 windows with approximately balanced class distribution. All experiments follow a leave-one-subject-out (LOSO) protocol, training on 50 subjects and evaluating on the held-out subject across all 51 folds.

\subsection{Bio-PM Encoder Pipeline}
Each window is encoded using two frozen Bio-PM modules. The accelerometer encoder applies a bandpass filter and projects the signal into 192 movement-element patches, each represented as a 64-dimensional token embedding via a bidirectional Transformer, producing a token sequence of shape $192\times64$. The gravity encoder applies a low-pass filter and maps the DC component to a 64-dimensional gravity embedding via a CNN, and appends it as a key to the same token\_store.h5. Neither encoder is fine-tuned at any stage so they serve exclusively as a frozen feature extractor and quality oracle for generation.

\subsection{Bio-PM Token Space Inspection}
\begin{figure}
    \centering
    \includegraphics[width=0.8\linewidth]{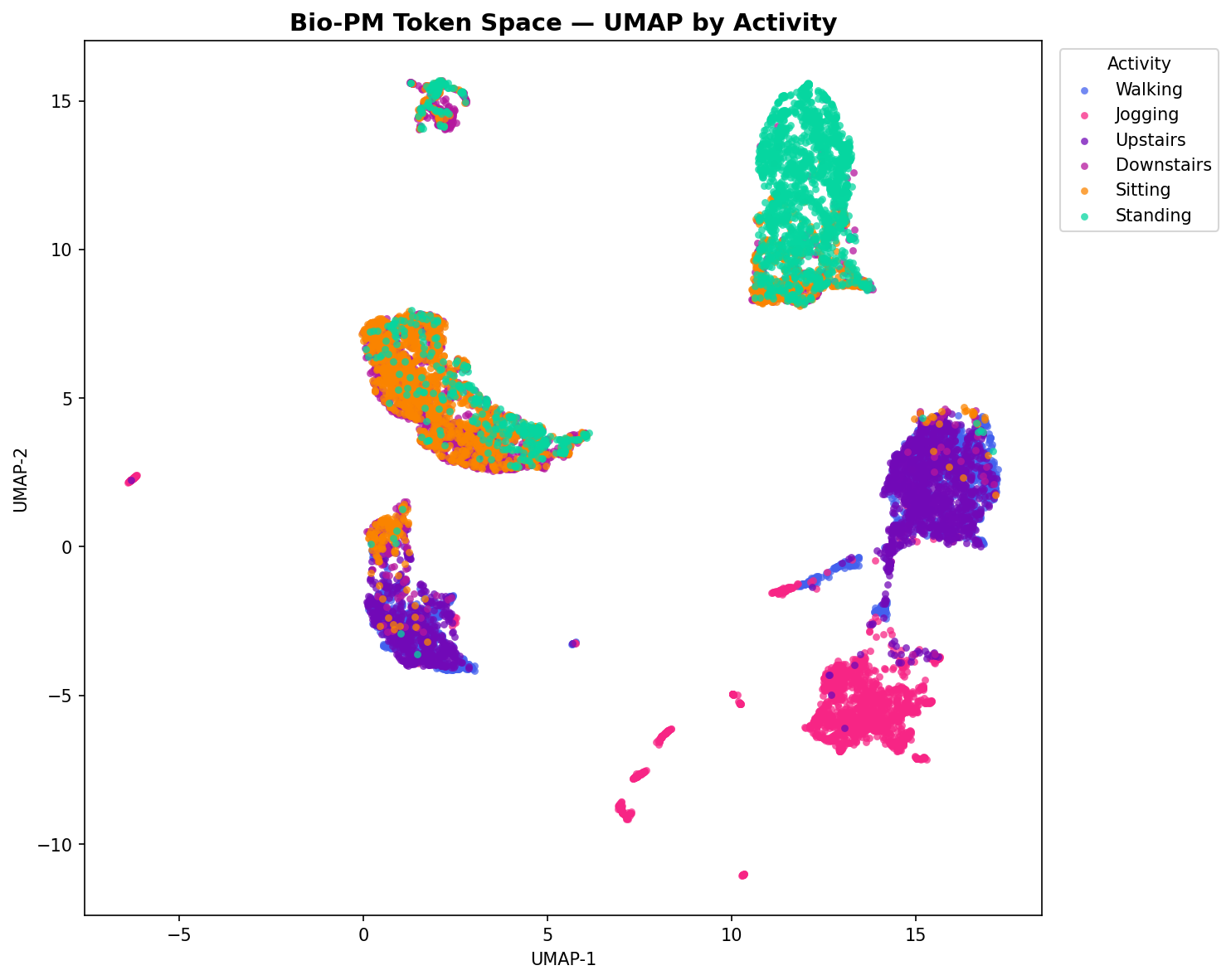}
    \caption{UMAP activity clusters on real data tokens generated by Bio-PM}
    \label{fig:umap_activity}
\end{figure}

To validate the token space before training any decoder, we compute nearest-neighbor same-class accuracy and visualize the mean-pooled token embeddings using UMAP. As shown in Fig.~\ref{fig:umap_activity}, activity classes form well-separated clusters with no subject-level structure. A nearest-neighbor classifier achieves 90\% same-class accuracy in this space, confirming semantic richness. The absence of subject separation means that a decoder conditioned only on activity class is sufficient, and no subject conditioning is required. These properties make Bio-PM's token space a viable target for class-conditional generation.
Despite approximate class balance ($\approx$16.6\% per class), the dataset exhibits significant representation overlap. Baseline analyses reveal high confusion rates between activities sharing fundamental motion patterns: Walking and Upstairs are confused 22\%--27\% of the time, while Downstairs is confused with Sitting and Standing up to 23\% of the time.

\section{CAR-IMU: Autoregressive Decoder}
\subsection{Architecture and Training}
CAR-IMU is a class-conditional autoregressive transformer that generates synthetic Bio-PM token sequences one position at a time. A learned activity-specific prefix token CLS\_c is prepended to the token sequence, providing class conditioning. The model consists of four causal Transformer layers with multi-head attention, where each position attends only to preceding positions. An output projection head maps each hidden state back to the 64-dimensional token space.
Training follows a next-token prediction objective using mean squared error (MSE) loss, with teacher forcing applied during training and autoregressive rollout at inference. The model is trained for 50 epochs using AdamW with a cosine learning rate schedule on the 12,514 real token sequences from the token store. The resulting model has 137K trainable parameters, with all Bio-PM encoder weights frozen throughout.

\subsection{The Calibration Problem}
Raw CAR-IMU tokens exhibit a distributional shift from real tokens. While centroid directions are roughly aligned, per-class means and variances differ measurably. We quantify generation realism using Train-on-Synthetic-Test-on-Real (TSTR) evaluation, where an MLP classifier is trained entirely on synthetic tokens and tested on real data. Uncalibrated CAR-IMU tokens yield a TSTR macro-F1 of 0.22, confirming the shift is large enough to mislead a downstream classifier.
To close this gap, we apply per-class affine calibration, shifting and scaling each synthetic class distribution to match the mean and standard deviation of the corresponding real class:

\begin{equation}
\hat{x} = \frac{x - \mu_{\text{syn},c}}{\sigma_{\text{syn},c}} \cdot \sigma_{\text{real},c} + \mu_{\text{real},c}
\end{equation}

After calibration, TSTR macro-F1 improves to 0.63 (more details in Table ~\ref{tab:decoder_quality}), demonstrating that the representational structure is present but distributionally offset. While calibration resolves the shift post-hoc, it requires access to real token statistics and does not address the underlying generation gap.

\subsection{Architectural Limitations}
While computationally efficient and highly effective for high-variance locomotion activities, the CAR-IMU architecture introduces structural challenges beyond distributional shift:
\begin{itemize}
    \item Class Signal Fading: Because the class embedding is only injected at position 0, the conditioning signal dilutes over the 192-step sequence.
    \item Autoregressive Drift: Prediction errors compound over time. This disproportionately affects static and less periodic activities (e.g., Sitting, Standing, Downstairs) that lack the self-correcting periodic rhythms naturally found in activities like Walking.
\end{itemize}

Together, these limitations, (1) distributional shift requiring post-hoc calibration, (2) class signal fading, and (3) autoregressive error compounding, motivate CD-IMU: a diffusion-based decoder that conditions on class identity at every denoising step and learns the full token distribution directly, without relying on calibration.

\section{CD-IMU: Class-Conditional Diffusion Decoder}

\subsection{Architecture}
CD-IMU is a class-conditional denoising diffusion model operating over Bio-PM token sequences. The forward process adds Gaussian noise to real token sequences over $T=1000$ timesteps using a linear schedule ($\beta_1 = 10^{-4}$, $\beta_T = 0.02$). The reverse process is parameterized by an 8-layer bidirectional Transformer that directly predicts the clean token sequence $\hat{x}_0$ from the noisy input $x_t$, the timestep $t$, and the activity class $c$.

Class and timestep conditioning are applied at every layer via Feature-wise Linear Modulation (FiLM) \cite{film}, modulating each layer's activations using scale and shift parameters from a concatenated class and sinusoidal timestep embedding. This ensures class conditioning persists throughout the full network depth, directly addressing the signal fading of CAR-IMU. Bidirectional attention with no causal mask is consistent with Bio-PM's contextual encoding.

The training objective combines three losses:
\begin{equation}
\mathcal{L} = \mathcal{L}_{x_0} + w(e)\left(\mathcal{L}_{\text{dist}} + \mathcal{L}_{\text{freq}}\right)
\end{equation}
where $\mathcal{L}_{x_0}$ is the MSE between predicted and real clean tokens, $\mathcal{L}_{\text{dist}}$ is a per-class mean and variance alignment term, $\mathcal{L}_{\text{freq}}$ penalizes spectral discrepancy along the sequence axis, and $w(e)$ is a linear warmup weight ramped from 0 to 1 over the first 15 epochs to prevent regularization from dominating before the model learns basic denoising. Generation uses deterministic DDIM sampling \cite{ddim} with 200 steps.

\subsection{Architectural Ablation}

An earlier variant introduced a Spectral FFN branch and an aggressive spectral loss coefficient $\beta_{\mathrm{freq}}{=}5.0$, causing the spectral loss to dominate reconstruction at a ratio of approximately $1369\!:\!1$ and preventing convergence. Additionally, checkpointing on validation loss caused the best checkpoint to be saved at epoch 1 due to a favorable initialization, yielding a model that generated class-agnostic tokens.

CD-IMU resolves these failures by reducing $\beta_{\mathrm{freq}}$ to $0.05$, removing the Spectral FFN, applying uniform regularization weights across classes, and checkpointing on average cosine similarity from periodic generation evaluations rather than validation loss.

\section{Classification Experiments}

\subsection{CAR-IMU Augmentation Strategies}

We evaluate CAR-IMU synthetic tokens across four augmentation experiments that vary calibration, feature dimensionality, and augmentation strategy. Features are extracted from the 192-token sequence in four configurations as summarised in Table~\ref{tab:c1c4}. The gravity embedding encodes device orientation and is particularly discriminative for postural activities such as Sitting and Standing, which are difficult to separate using accelerometer statistics alone.

\begin{table}[htbp]
\caption{CAR-IMU Augmentation Experiment Configurations}
\label{tab:c1c4}
\centering
\begin{tabular}{cllll}
\toprule
\textbf{Exp.} & \textbf{Calib.} & \textbf{Aug. Strategy} & \textbf{Dim} & \textbf{Features} \\
\midrule
Exp1 & No  & 5\% uniform      & 64-d  & $\mu_{\mathrm{acc}}$ \\
Exp2 & Yes & 5\% uniform      & 64-d  & $\mu_{\mathrm{acc}}$ \\
Exp3 & Yes & 5\% uniform      & 192-d & $\mu_{\mathrm{acc}},\ \sigma_{\mathrm{acc}},\ \mathbf{g}$ \\
Exp4 & Yes & Difficulty-aware & 320-d & $\mu_{x},\ \mu_{y},\ \mu_{z},\ \sigma_{\mathrm{acc}},\ \mathbf{g}$ \\
\bottomrule
\end{tabular}
\end{table}

Exp1 uses uncalibrated tokens to establish the cost of distributional shift. Exp2 introduces per-class affine calibration. Exp3 extends Exp2 with a richer feature space. Exp4 applies a difficulty-aware augmentation strategy, assigning higher synthetic ratios to classes frequently misclassified in the baseline LOSO confusion matrix, while withholding augmentation from classes already near ceiling performance. All of these are tested against only-real data baselines. 

For each experiment, we explored three classifier architectures: logistic regression, a three-layer MLP, and an MLP with an attention pooling layer over the token sequence. Final results for CAR-IMU augmentations are reported for the MLP with attention pooling, which consistently outperformed the alternatives on held-out validation folds. All experiments follow the LOSO protocol with a \texttt{StandardScaler} fitted on the training split of each fold.

\subsection{CD-IMU Classification Pipeline}
CD-IMU tokens are evaluated under two strategies. As seen in Table~\ref{tab:e1e2}, Exp1 is a vanilla baseline with uniform 5\% augmentation across all classes using 64-d features. Exp2 uses confusion-aware augmentation with 320-d features, where per-class synthetic ratios are derived from the baseline LOSO confusion matrix. Classes identified as most frequently misclassified (Sitting, Downstairs) receive higher augmentation ratios. Jogging, which is near the performance ceiling in the baseline, receives no augmentation. All of these are tested against only-real data baselines. 

\begin{table}[htbp]
\caption{CD-IMU Augmentation Experiment Configurations}
\label{tab:e1e2}
\centering
\begin{tabular}{cllll}
\toprule
\textbf{Exp.} & \textbf{Calib.} & \textbf{Aug. Strategy} & \textbf{Dim} & \textbf{Features} \\
\midrule
Exp1 & No  & 5\% uniform      & 64-d  & $\mu_{\mathrm{acc}}$ \\
Exp2 & No & Difficulty-aware & 320-d & $\mu_{x},\ \mu_{y},\ \mu_{z},\ \sigma_{\mathrm{acc}},\ \mathbf{g}$ \\
\bottomrule
\end{tabular}
\end{table}

As with the CAR-IMU experiments, logistic regression, MLP, and MLP with attention pooling were evaluated. Final results are reported for the standard MLP, which generalised best on the higher-dimensional 320-d feature space. The same LOSO protocol and per-fold scaling are applied throughout.

\section{Reconstruction Pipeline}

The reconstruction pipeline maps synthetic Bio-PM token sequences back to approximate raw IMU waveforms. Rather than reconstructing at the window level, we exploit the direct correspondence between each Bio-PM token and the movement-element patch it was computed from, treating reconstruction as a per-token regression problem.

\subsection{PatchDecoder Architecture}
PatchDecoder is a shared-backbone MLP applied independently to each of the 192 tokens in a sequence. Given a 64-dimensional token, a three-layer backbone with GELU activations produces a 256-dimensional hidden representation, which is passed to four task-specific heads: (1) a waveform head predicting the 32-sample normalized patch signal; (2) an axis head classifying the accelerometer axis ($X$, $Y$, or $Z$); (3) a position head predicting the patch's temporal location within the window; and (4) a duration head predicting the original patch length before normalization. The model is trained on approximately 1.4M patch-level pairs using a composite loss combining MSE and cosine similarity on the waveform, cross-entropy on the axis, and MSE on position and duration.

\subsection{Waveform Stitching}
At inference, each predicted 32-sample patch is resampled to its predicted duration via cubic interpolation, placed at its predicted position on its predicted axis in a $(300 \times 3)$ output buffer, and averaged where patches overlap. A per-class gravity DC offset derived from real data is added to restore low-frequency orientation content. The final output is a $(300 \times 3)$ approximate IMU waveform per synthetic token sequence.

\subsection{Limitations}
Because each token is decoded independently without access to neighboring tokens, the decoder cannot enforce temporal continuity across patch boundaries, resulting in discontinuities in the stitched waveform. Amplitude attenuation is also observed, as the per-token regression averages overlapping patches rather than preserving peak dynamics. Addressing these limitations through a context-aware decoder and overlap-add stitching is left as future work.

\section{Evaluation}

\subsection{Decoder Results}
Table~\ref{tab:decoder_quality} compares token generation quality across CAR-IMU (uncalibrated), CAR-IMU (calibrated), and CD-IMU. Calibration substantially improves CAR-IMU's TSTR score, confirming that the representational structure is present but distributionally offset. CD-IMU achieves a higher TSTR than calibrated CAR-IMU without any post-hoc correction, demonstrating that the diffusion objective learns the token distribution directly. Cosine similarity for calibrated CAR-IMU is 1.0 by construction, since calibration matches each class mean, so it is not informative for that column. The near-unity cosine similarity and low FD scores of CD-IMU across all six activity classes confirm that CD-IMU synthetic tokens are distributionally close to real tokens in Bio-PM's representation space.

\begin{figure*}[htbp]
    \centering
    \hspace*{0.05\linewidth}%
    \begin{subfigure}[t]{0.40\linewidth}
        \centering
        \includegraphics[width=\linewidth,height=4.6cm]{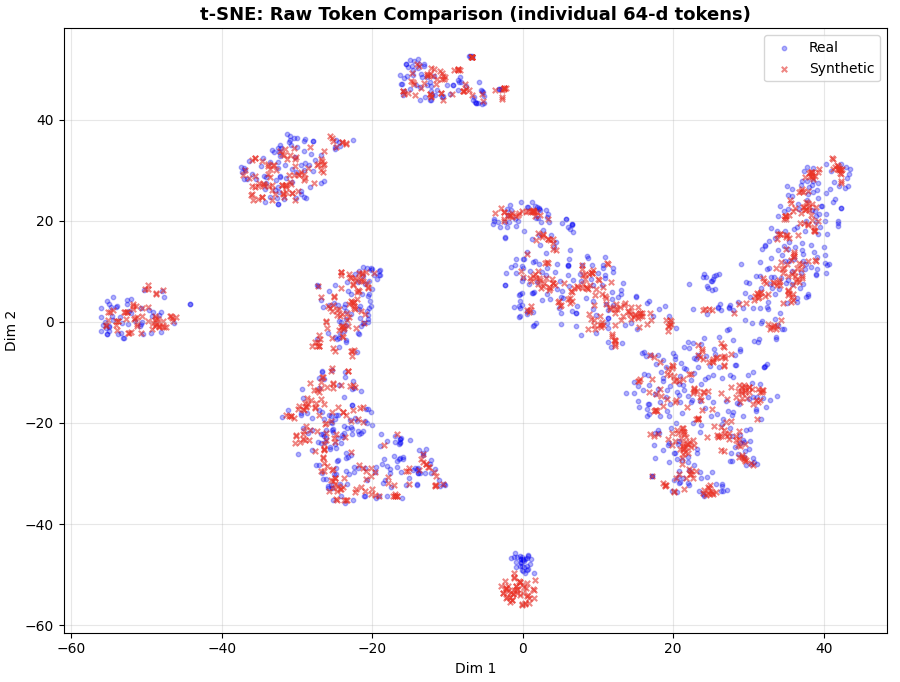}
        \caption{t-SNE: Raw Tokens}
        \label{fig:tsne_raw}
    \end{subfigure}%
    \hfill%
    \begin{subfigure}[t]{0.40\linewidth}
        \centering
        \includegraphics[width=\linewidth,height=4.6cm]{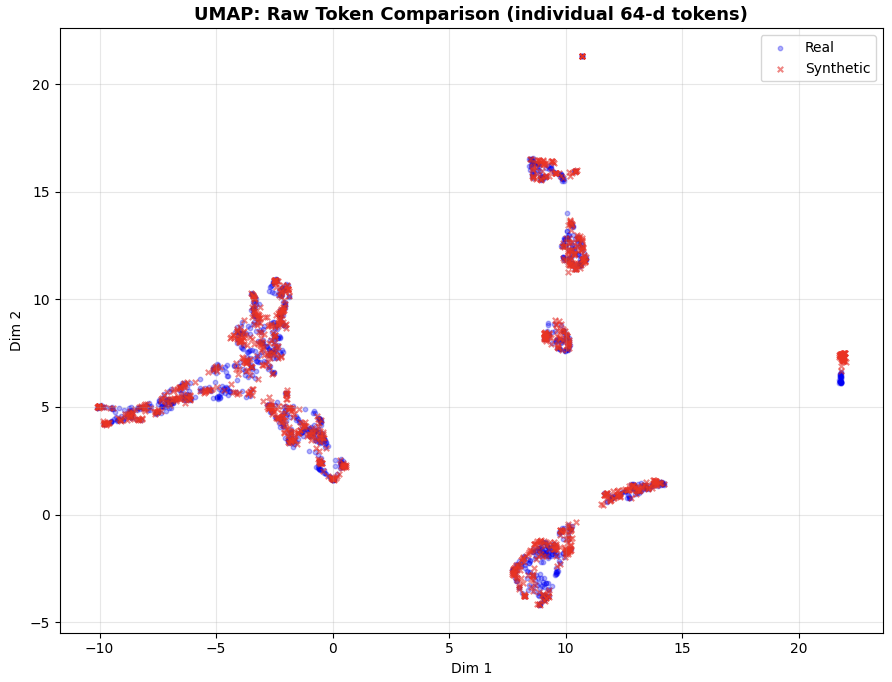}
        \caption{UMAP: Raw Tokens}
        \label{fig:umap_raw}
    \end{subfigure}%
    \hspace*{0.05\linewidth}%

    \vspace{0.3cm} %

    \hspace*{0.05\linewidth}%
    \begin{subfigure}[t]{0.40\linewidth}
        \centering
        \includegraphics[width=\linewidth,height=4.6cm]{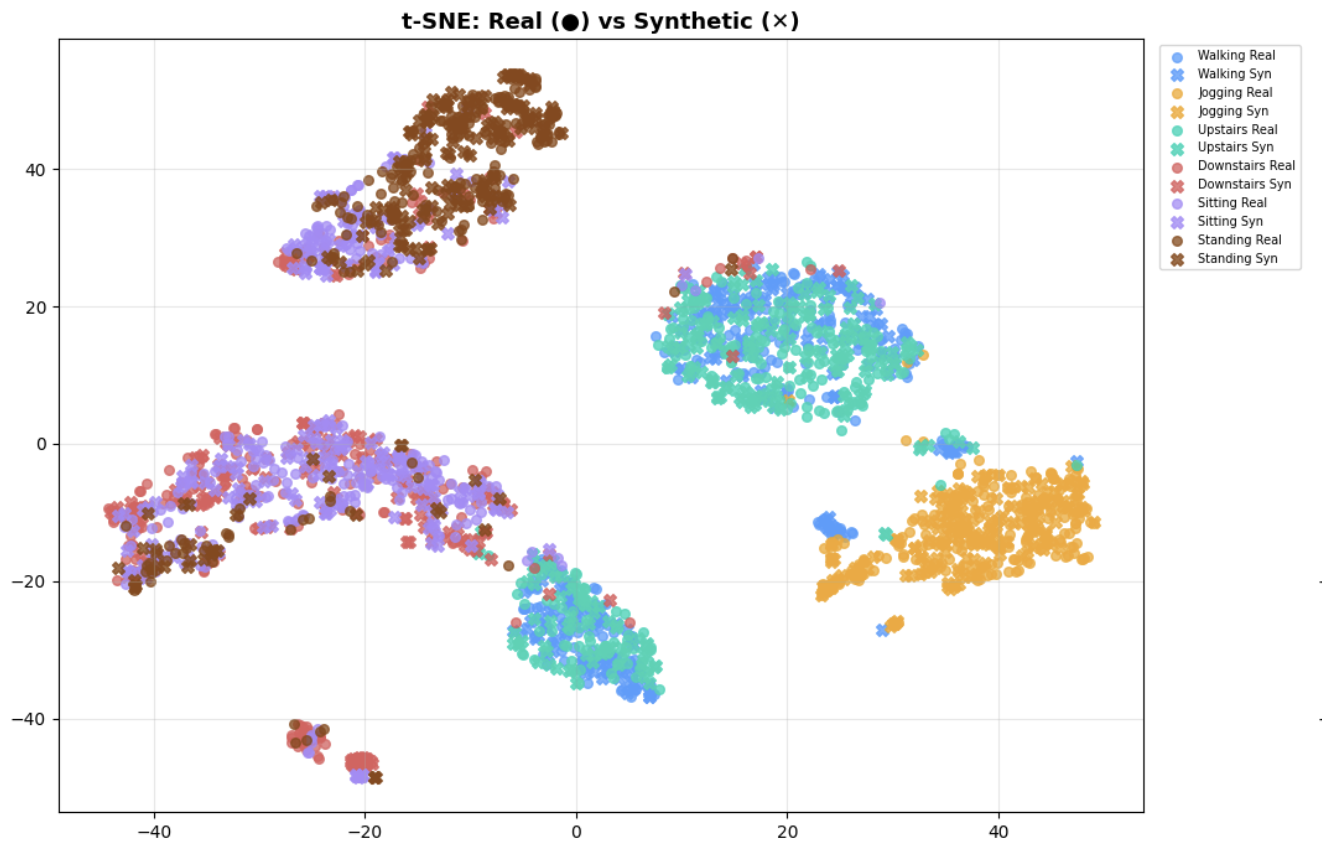}
        \caption{t-SNE: Mean-Pooled}
        \label{fig:tsne_pooled}
    \end{subfigure}%
    \hfill%
    \begin{subfigure}[t]{0.40\linewidth}
        \centering
        \includegraphics[width=\linewidth,height=4.6cm]{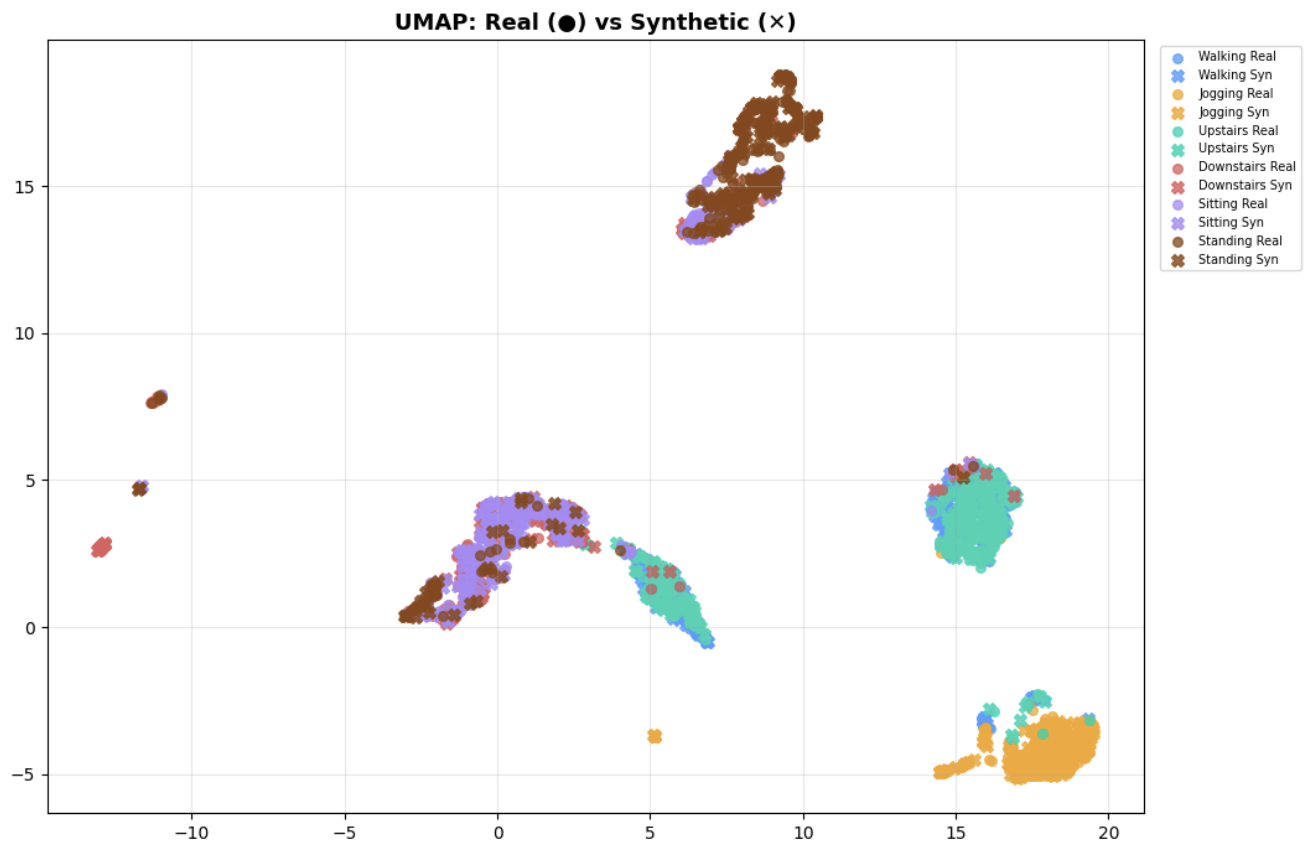}
        \caption{UMAP: Mean-Pooled}
        \label{fig:umap_pooled}
    \end{subfigure}%
    \hspace*{0.05\linewidth}%

    \caption{Embedding space comparison of real and CD-IMU synthetic tokens at two levels of granularity. 
(a, b) Raw individual 64-d token embeddings sampled across all sequence positions: 
real (blue circles) and synthetic (red crosses) are intermixed throughout all clusters, 
confirming distributional alignment at the patch level. 
(c, d) Mean-pooled window-level features colored by activity class: 
synthetic tokens for each class co-locate with their real counterparts, 
preserving the semantic structure of Bio-PM's token space. 
Together, (a--d) demonstrate that CD-IMU generates tokens that are 
realistic both at the individual patch position level and at the 
window-level activity representation level.}
    \label{fig:tsne_decoder}
\end{figure*}

Fig.~\ref{fig:tsne_decoder} shows t-SNE and UMAP projections of real and CD-IMU synthetic tokens. Synthetic tokens for each class are tightly co-located with their real counterparts, and the raw token-level visualization confirms alignment at the individual patch position level rather than only at the window mean.

\subsection{Classification: CAR-IMU}

Table~\ref{tab:car_clf} reports LOSO macro-F1 across the four CAR-IMU augmentation experiments. Exp1 confirms that uncalibrated tokens degrade classifier performance relative to the real-only baseline. Calibration (Exp2) recovers a modest positive gain. Extending to richer 192-d features (Exp3) and confusion-aware augmentation with 320-d features (Exp4) yields the best CAR-IMU result. All experiments use MLP with attention pooling as the final classifier.

\begin{table}[htbp]
\caption{LOSO Classification Results: CAR-IMU Augmentation Experiments}
\label{tab:car_clf}
\centering
\resizebox{\columnwidth}{!}{%
\begin{tabular}{llrrr}
\toprule
\textbf{Exp.} & \textbf{Features} & \textbf{Real Macro-F1} & \textbf{Real+Syn Macro-F1} & \textbf{$\Delta$} \\
\midrule
CAR-Exp1  & 64-d  &  0.796 & 0.790 & $-0.006$ \\
CAR-Exp2  & 64-d  &  0.787 & 0.790 & $+0.003$ \\
CAR-Exp3  & 192-d &  0.798 & 0.805 & $+0.007$ \\
CAR-Exp4  & 320-d &  0.791 & 0.801 & $+0.010$ \\
\bottomrule
\end{tabular}%
}
\end{table}

\subsection{Classification: CD-IMU}

\begin{figure}[htbp]
\centerline{\includegraphics[width=0.85\columnwidth]{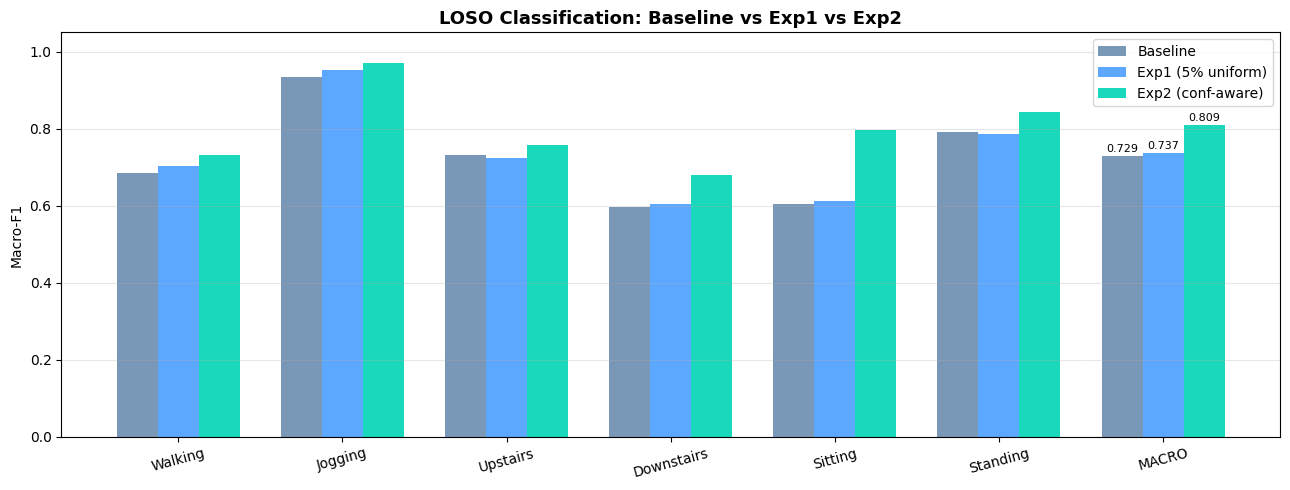}}
\caption{Per-class and macro-average F1 for Baseline, CD-Exp1, and CD-Exp2 across 51 LOSO folds on WISDM. CD-Exp2 consistently improves F1 across all activity classes.}
\label{fig:clf_bar}
\end{figure}

Table~\ref{tab:cdm_clf} reports results for Exp1 and Exp2. Exp1's marginal and statistically insignificant gain confirms that token quality alone is insufficient without richer feature representation. Exp2's statistically significant improvement of $+0.080$ macro-F1 ($p < 0.001$) demonstrates the combined effect of 320-d features, gravity encoding, and confusion-aware augmentation. Since the baseline uses 64-d features, this gain reflects the full pipeline; at matched 64-d features, augmentation contributes $+0.008$ (Exp1). The Sitting class shows the largest per-class improvement at $+0.191$, driven by the gravity embedding's discriminative power for postural activities. Fig.~\ref{fig:clf_bar} summarises per-class F1 across all three conditions.

\begin{table}[htbp]
\caption{LOSO Classification Results: CD-IMU Augmentation Experiments}
\label{tab:cdm_clf}
\centering
\begin{tabular}{llrrrr}
\toprule
\textbf{Exp.} & \textbf{Features} & \textbf{Macro-F1} & \textbf{$\Delta$} & \textbf{$p$-value} \\
\midrule
Baseline & 64-d  & 0.7288 & --     & --     \\
CD-Exp1       & 64-d  & 0.7369 & +0.008 & 0.101  \\
CD-Exp2       & 320-d & 0.8091 & +0.080 & $<$0.001 \\
\bottomrule
\end{tabular}
\end{table}

\subsection{Reconstruction Results}
\begin{figure}[htbp]
\centerline{\includegraphics[width=\columnwidth]{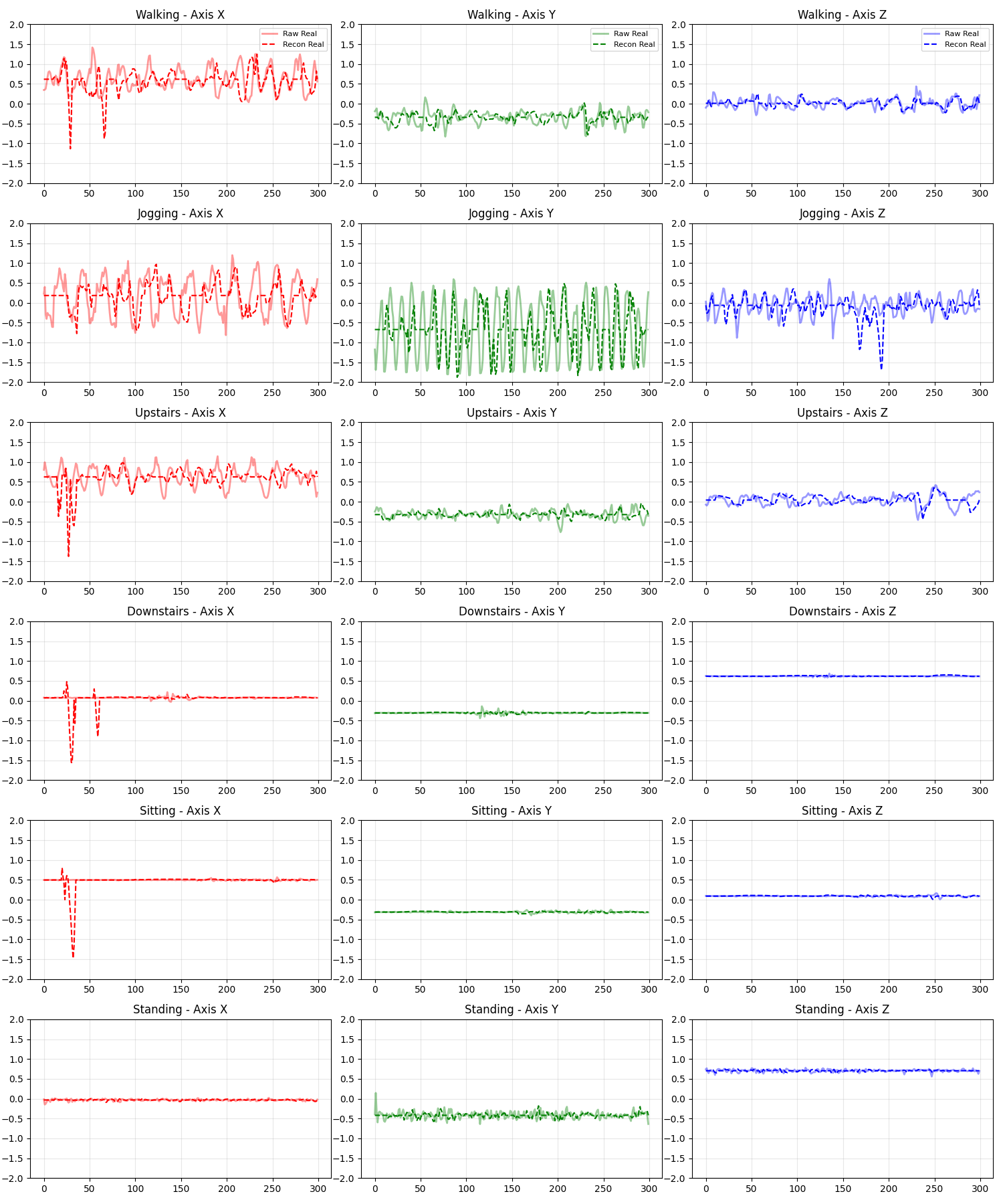}}
\caption{PatchDecoder reconstruction: real waveform (solid) vs reconstructed (dashed) for activity classes across X, Y, Z axes. Frequency content and temporal rhythm are well-preserved: amplitude 
attenuation is visible for high-variance locomotion classes.}
\label{fig:reconstruction_1}
\end{figure}

\begin{figure}[htbp]
\centerline{\includegraphics[width=\columnwidth]{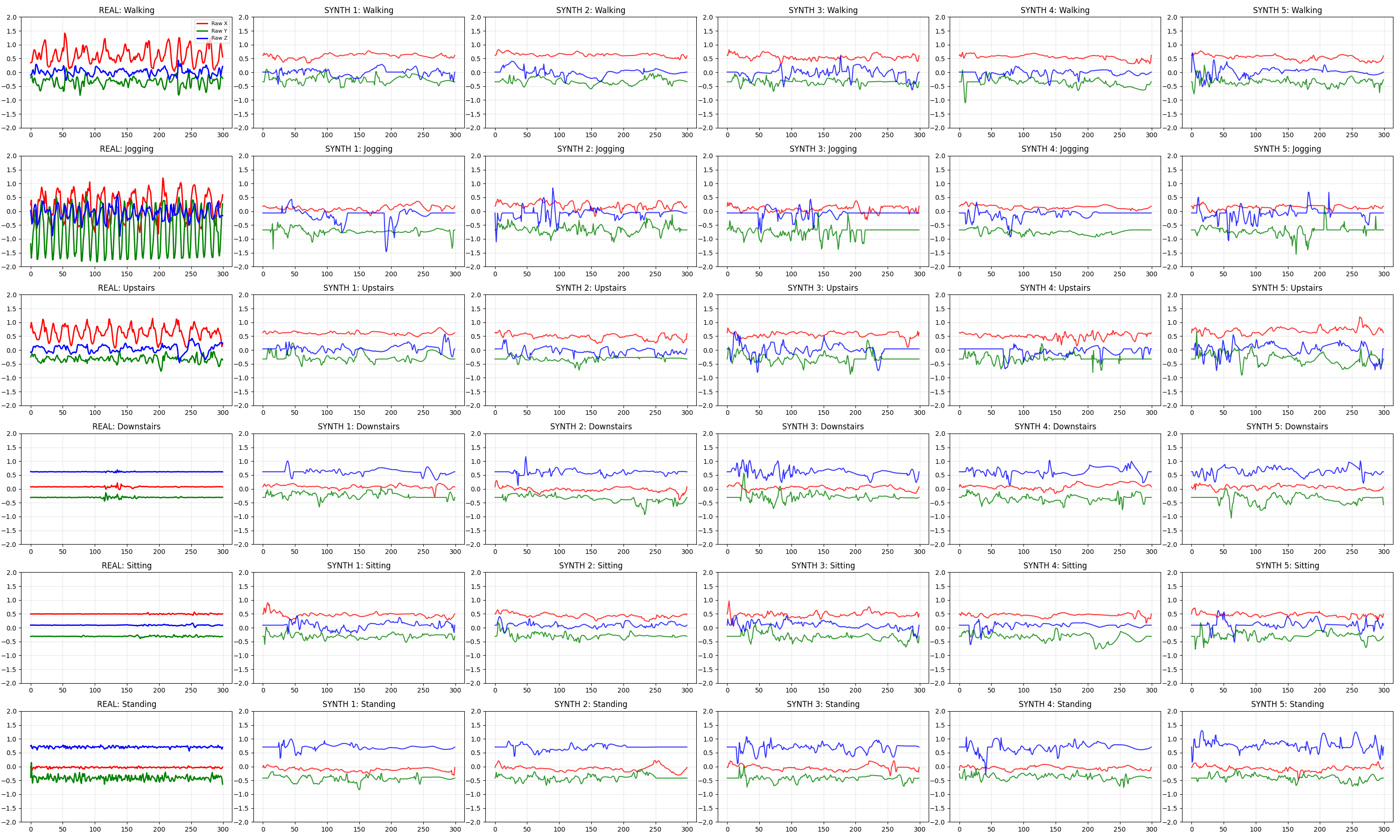}}
\caption{Gallery of CAR-IMU synthetic waveforms decoded via PatchDecoder (columns 2--6) 
alongside the real reference waveform (column 1) for each activity class. Static activities (Sitting, Standing) and Downstairs are reconstructed with 
faithful amplitude and DC offset. Locomotion classes (Walking, Jogging, Upstairs) exhibit correct oscillation frequency but attenuated amplitude, 
consistent with the per-patch independence of the decoder.}
\label{fig:reconstruction_2}
\end{figure}

PatchDecoder reconstruction is evaluated against raw IMU waveforms using Spectral Correlation, Zero-Crossing rate (ZCR) error, Signal Magnitude Area (SMA) error, and waveform RMSE. Spectral correlation exceeds 0.98 for five of six classes, and ZCR error remains below 0.07 across all activities, indicating that the frequency content and temporal rhythm of real waveforms are well-preserved in reconstruction. However, SMA error is substantially elevated for locomotion activities like 0.56 for Walking and 0.96 for Jogging, revealing a consistent amplitude attenuation in the decoded waveforms. The alignment of real reconstructed waveform on top of real waveform is visible in Fig.~\ref{fig:reconstruction_1}. While the reconstructed signals capture the correct oscillation frequency and axis assignment, their amplitude is significantly lower than real waveforms, particularly for high-variance activities such as Jogging as seen in Fig.~\ref{fig:reconstruction_2}. The attenuation arises from the per-patch independence assumption: the decoder processes each token without awareness of neighbouring patches, and the overlap-add averaging step further smooths out amplitude peaks. The low waveform RMSE (0.001--0.010) is misleading in this context, as it reflects the absence of large absolute errors rather than faithful amplitude reproduction.

\subsection{Discussion}
CD-IMU eliminates the calibration dependency of CAR-IMU by learning the token distribution directly through a denoising objective with FiLM conditioning at every layer.
The step from CD-Exp1 to CD-Exp2 highlights that token quality is necessary but not sufficient, so richer feature representations and targeted augmentation strategies are equally important for downstream gains. Sitting's improvement of $+0.191$ underscores the value of incorporating the gravity embedding, which encodes device orientation and sharply distinguishes postural activities that are otherwise ambiguous in accelerometer statistics alone. The PatchDecoder demonstrates feasibility of waveform reconstruction from token sequences but is constrained by patch-level independence. Context-aware reconstruction decoding with overlap-add stitching remains an open direction.

\begin{table*}[htbp]
\caption{Token Generation Quality per Activity: CAR-IMU vs CD-IMU}
\label{tab:decoder_quality}
\centering
\resizebox{\textwidth}{!}{%
\begin{tabular}{l ccc ccc ccc ccc ccc}
\toprule
\multirow{2}{*}{\textbf{Activity}} & \multicolumn{3}{c}{\textbf{Cos Sim $\uparrow$}} & \multicolumn{3}{c}{\textbf{FD $\downarrow$}} & \multicolumn{3}{c}{\textbf{MMD $\downarrow$}} & \multicolumn{3}{c}{\textbf{Var Ratio}} & \multicolumn{3}{c}{\textbf{TSTR F1 $\uparrow$}} \\
\cmidrule(lr){2-4} \cmidrule(lr){5-7} \cmidrule(lr){8-10} \cmidrule(lr){11-13} \cmidrule(lr){14-16}
& \textbf{CAR (uncal)} & \textbf{CAR (cal)} & \textbf{CD} & \textbf{CAR (uncal)} & \textbf{CAR (cal)} & \textbf{CD} & \textbf{CAR (uncal)} & \textbf{CAR (cal)} & \textbf{CD} & \textbf{CAR (uncal)} & \textbf{CAR (cal)} & \textbf{CD} & \textbf{CAR (uncal)} & \textbf{CAR (cal)} & \textbf{CD} \\
\midrule
Walking    & 0.8433 & 1.0000 & 0.9995 & 24.95 & 10.68 & 0.17 & 0.1505 & 0.1539 & 0.0206 & 2.933 & 2.412 & 0.947 & 0.00 & 0.48 & 0.67 \\
Jogging    & 0.7236 & 1.0000 & 0.9983 & 26.70 &  7.93 & 0.31 & 0.2355 & 0.2033 & 0.0524 & 2.021 & 2.239 & 0.960 & 0.02 & 0.90 & 0.95 \\
Upstairs   & 0.8515 & 1.0000 & 0.9991 & 23.49 & 10.64 & 0.19 & 0.1210 & 0.1217 & 0.0214 & 3.108 & 2.621 & 0.941 & 0.49 & 0.54 & 0.63 \\
Downstairs & 0.7253 & 1.0000 & 0.9966 & 30.75 &  6.50 & 0.70 & 0.0338 & 0.0375 & 0.0140 & 1.381 & 1.086 & 1.005 & 0.21 & 0.50 & 0.53 \\
Sitting    & 0.7770 & 1.0000 & 0.9984 & 26.95 &  5.74 & 0.42 & 0.0639 & 0.0604 & 0.0178 & 1.707 & 1.320 & 0.985 & 0.07 & 0.57 & 0.56 \\
Standing   & 0.8763 & 1.0000 & 0.9984 & 23.49 &  8.44 & 0.28 & 0.2333 & 0.2339 & 0.0359 & 1.784 & 1.335 & 0.986 & 0.50 & 0.76 & 0.71 \\
\midrule
\textbf{AVERAGE} & \textbf{0.7995} & \textbf{1.0000} & \textbf{0.9984} & \textbf{26.06} & \textbf{8.32} & \textbf{0.35} & \textbf{0.1397} & \textbf{0.1351} & \textbf{0.0270} & \textbf{2.156} & \textbf{1.836} & \textbf{0.971} & \textbf{0.2152} & \textbf{0.6254} & \textbf{0.6755} \\
\bottomrule
\end{tabular}%
}
\end{table*}

\section{Conclusion}

We presented a pipeline for generating synthetic IMU representations in the token embedding space of Bio-PM, using the frozen pretrained encoder as a quality oracle. Two class-conditional decoders were developed and systematically compared: CAR-IMU, an autoregressive transformer that generates token sequences one position at a time, and CD-IMU, a diffusion decoder conditioned at every layer via FiLM modulation. CAR-IMU produces structurally plausible tokens but requires per-class affine calibration to close the distributional gap, a dependency that reveals the limits of autoregressive generation in contextual representation spaces. CD-IMU eliminates this requirement by learning the full token distribution through a denoising objective, achieving a TSTR macro-F1 of 0.68 without calibration and outperforming the calibrated CAR-IMU baseline.
Augmenting a LOSO HAR classifier with CD-IMU synthetic tokens, combined with 320-dimensional features incorporating the gravity embedding and a confusion-aware augmentation strategy, improves macro-F1 from 0.729 to 0.809 across 51 subjects on the WISDM benchmark, a statistically significant gain of 0.080 points. The Sitting activity class improves by 0.191, driven by the gravity embedding's discriminative power for postural activities. PatchDecoder reconstruction demonstrates that synthetic token sequences can be approximately mapped back to raw IMU waveforms, with strong spectral fidelity but persistent amplitude attenuation for high-variance locomotion classes, motivating context-aware decoding as future work.

\small
\setlength{\itemsep}{0pt}

\end{document}